\documentclass{article}

\usepackage[preprint]{neurips_2023}

\usepackage[utf8]{inputenc} 
\usepackage[T1]{fontenc}    
\usepackage{hyperref}       
\usepackage{url}            
\usepackage{booktabs}       
\usepackage{amsfonts}       
\usepackage{nicefrac}       
\usepackage{microtype}      
\usepackage{xcolor}         

\usepackage{bm}             
\usepackage{amsmath}        
\usepackage{svg}
\usepackage{graphicx}
\graphicspath{{graphics}}
\usepackage{tikz}           
\usetikzlibrary{trees}
\usepackage{caption}
\usepackage{subcaption}
\usepackage{booktabs}
\usepackage{pifont}
\usepackage{array} 

\usepackage{lineno}         

\title{Reducing Annotation Burden in Physical Activity Research Using Vision-Language Models}

\author{%
Abram Schönfeldt\\
Department of Population Health, University of Oxford, Oxford, United Kingdom\\
\And
Benjamin Maylor\\
Department of Population Health, University of Oxford, Oxford, United Kingdom\\
\And
Xiaofang Chen\\
School of Epidemiology and Health Statistics, Chengdu Medical College, Sichuan, China\\
\And
Ronald Clark \\
Department of Computer Science, University of Oxford, Oxford, United Kingdom\\
\And
Aiden Doherty   \thanks{Corresponding author: \texttt{aiden.doherty@ndph.ox.ac.uk}}\\
Department of Population Health, University of Oxford, Oxford, United Kingdom\\
}

\begin{document}

\maketitle

\begin{abstract}
\textit{Introduction:} Data from wearable devices collected in free-living settings, and labelled with physical activity behaviours compatible with health research, are essential for both validating existing wearable-based measurement approaches and developing novel machine learning approaches. 
One common way of obtaining these labels relies on laborious annotation of sequences of images captured by cameras worn by participants through the course of a day. Open-source vision language models, which can be run locally, could be prompted to predict physical activity behaviours, reducing the burden of human annotation.
\textit{Methods:} We compare the performance of three vision language models and two discriminative models on two free-living validation studies with 161 and 111 participants, collected in Oxfordshire, United Kingdom and Sichuan, China, respectively, using the Autographer (OMG Life, defunct) wearable camera.
\textit{Results:} We found that the best open-source vision-language model (VLM) and fine-tuned discriminative model (DM) achieved comparable performance when predicting sedentary behaviour from single images on unseen participants in the Oxfordshire study; median F$_1$-scores: VLM = 0.89 (0.84, 0.92), DM = 0.91 (0.86, 0.95). Performance declined for light (VLM = 0.60 (0.56,0.67), DM = 0.70 (0.63, 0.79)), and moderate-to-vigorous intensity physical activity (VLM = 0.66 (0.53, 0.85); DM = 0.72 (0.58, 0.84)). When applied to the external Sichuan study, performance fell across all intensity categories, with median Cohen's $\kappa$ scores falling from 0.54 (0.49, 0.64) to 0.26 (0.15, 0.37)  for the VLM, and from 0.67 (0.60, 0.74) to 0.19 (0.10, 0.30) for the DM.  
\textit{Conclusion:} Freely available computer vision models\footnote{Code will be made available at \url{https://github.com/oxwearables}.} could help annotate sedentary behaviour, typically the most prevalent activity of daily living, from wearable camera images within similar populations to seen data, reducing the annotation burden. 
\end{abstract}

\begin{figure}
    \centering
    \includegraphics[width=\linewidth]{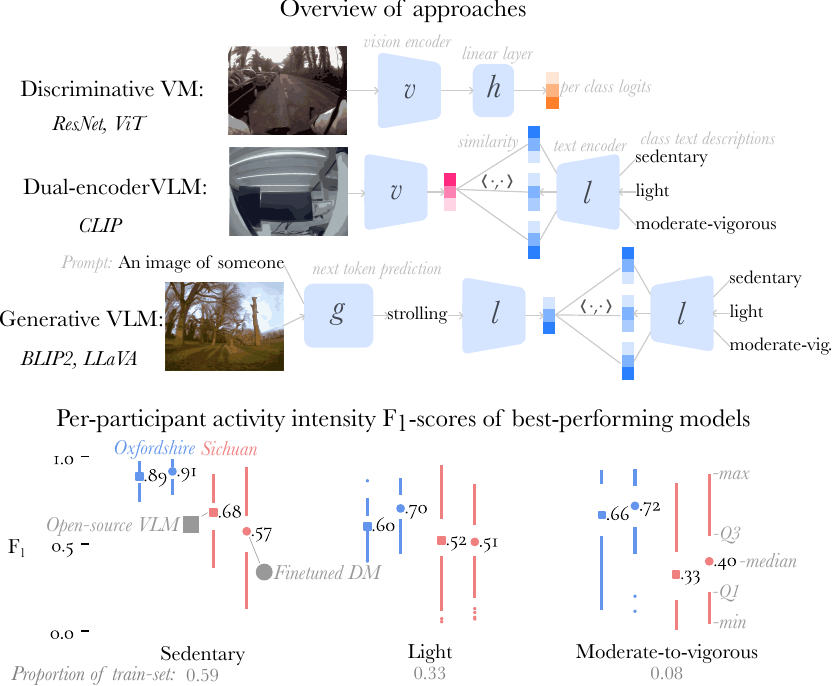}
    \caption{Illustration of the computer vision approaches compared (top). Below, quartile plots \citep{tufte_visual_2002} show the five-number summary of per-participant F$_1$-scores for Sedentary Behaviour (SB), Light Intensity Physical Activity (LIPA), and Moderate-to-Vigorous Physical Activity (MVPA), for the best-performing vision-language model, LLaVA (squares), and the best-performing discriminative vision model, ViT (circles), selected via hyperparameter tuning. Performance is shown for participants in the Oxfordshire study (blue) and the Sichuan study (red) withheld from model selection. MVPA constitutes only 8\% of the training set, which is reflected in the high variance of per-participant F$_1$-scores.}
    \label{fig:main}
\end{figure}

\section{Introduction}

Wearable measurements of physical activity behaviours have helped advance our understanding of the relationship between physical activity and health outcomes \citep{wasfy2022examining}, provided more sensitive outcomes in clinical trials
\citep{servais2023first} and introduced new ways of monitoring population physical activity levels \citep{troiano2020can}. The most realistic setting for validating behaviour measurement approaches and developing novel machine learning approaches \citep{logacjov_selfpab_2024, yuan2024self, walmsley2022reallocation, willetts2018statistical, doherty2017large} is in diverse populations of people living their everyday lives, highlighting the need for large, labelled, wearable data-sets, captured in free-living conditions \citep{bao2004activity,keadle2019framework, thomaz2023acquisition}.

Activity intensity classes, Sedentary Behaviour (SB), Light Intensity Physical Activity (LIPA) and Moderate-to-Vigorous Physical Activity (MVPA), provide a simple classification of daily activities based on their energy expenditure, are clearly defined \citep{tremblay2017sedentary, ainsworth2011compendium, keadle2024using}, and have been widely adopted in epidemiological research \citep{walmsley2022reallocation, schalkamp_wearable_2023, shreves_dose-response_2023} and physical activity guidelines \citep{bull_world_2020}. Although the gold standard for measuring activity intensity is video-recorded direct observation \citep{keadle2019framework}, which involves having participants followed by researchers filming their activities, a pragmatic approach to collecting these data-sets in free-living settings has been for the participants to wear cameras, which record footage that later is reviewed by annotators to inform the ground-truth labels \citep{chan2024capture, thomaz2023acquisition}. However, the sensitive nature of this footage has meant that access to it is restricted to select researchers, trained to handle sensitive data \cite{kelly2013ethical}, making it costly and time-consuming to label.

Recently, \cite{keadle2024using} proposed adopting approaches from computer vision to predict aspects of physical activity in a study of 26 adults, using video-recorded direct observation, emphasising the distinction between the definitions of physical activity used in health research \citep{tremblay2017sedentary, ainsworth2024brief, ATUS2024}, such as activity intensity, and the varied definitions of activity prevalent in human activity recognition literature \cite{herath2017going}. This work  estimates the performance of computer vision methods based on video-recorded direct observation, leaving the performance on studies using wearable image-capturing cameras unexplored, in addition to questions of how stable model performance will be between different populations, and within larger populations.

In this work, we evaluate activity intensity prediction using open-source Vision Language Models (VLMs), and Discriminative Models (DMs) on two validation studies collected in Oxfordshire, United Kingdom \citep{chan2024capture} and Sichuan, China \citep{chen2023device}, with wearable camera data from 161 and 111 participants respectively. Although ethical issues prevent us from making the wearable camera portion of these data-set publicly available, a detailed quality assessment of these data-sets is conducted, and we will make our codebase and models publicly available (the annotated wrist-worn accelerometer data is publicly available for the Oxfordshire study \citep{chan2024capture}). To our knowledge, this is the first work which assesses activity intensity prediction from wearable cameras using these methods.

\section{Relevant work}

\subsection{Wearable data-sets of health-relevant behaviours}

\begin{table}
    \centering
    \caption{Number of participants and estimated number of labelled hours of studies using cameras to validate wearable measurements of physical activity identified in a recent systematic review \citep{giurgiu2022quality}, and scoping review \citep{martinez2024accuracy}.}
    \begin{tabular}{ll>{\raggedright\arraybackslash}p{0.1\linewidth}l>{\raggedright\arraybackslash}p{0.1\linewidth}}
        \hline
        Paper&Viewpoint& No. participants &Median $\delta t$ (s)& Hours labelled \\
        \hline
        \cite{chan2024capture} (Oxfordshire) & 1st& 161 &24&1 546\\
        \cite{chen2023device} (Sichuan) & 1st& 111 &84&1 078\\
        \hline
        \cite{chasan_taber_update_2023} &1st& 50  &15& 1 218\\ 
        \cite{femiano_validation_2022} & 1st & 22  &Video& 11 \\ 
        \cite{van_alphen_construct_2021} &3rd& 22  &Video& 34.3 \\ 
        \cite{nawab_2021} &1st& 25  &20& 768 \\ 
        \cite{bach_machine_2021} &1st& 22  &Video& 38 \\ 
        \cite{marcotte_estimating_2020} &3rd& 48  &Video & 192 \\ 
        \cite{koenders_validation_2018} &3rd& 31  &Video& 31 \\ 
        \hline
    \end{tabular}

    \vspace{0.5em}
    \footnotesize
     \textit{Note:} this table is not an exhaustive, and we recommend referring to the reviews for a more comprehensive list of validation studies. The two studies collected in Oxfordshire and Sichuan used in this work are shown at the top of this table. The estimates of the number of hours of labelled data for the timelapse studies is optimistic, since the temporal resolution of the is much lower than video, resulting in periods of time that are difficult to label.
    \label{tab:validation_studies}
\end{table}

There are varying approaches to capturing free-living data-sets using cameras, arising from where the cameras are positioned relative to the participants, and the frequency with which cameras capture frames. Cameras can be worn by the participants, resulting in \textit{egocentric} footage, held by observers following the participants, or placed in static positions, with the latter two options resulting in \textit{third-person} footage. The frame-rate can be high, as is the case with video, or low, resulting in sparse sequences of images, similar to a time-lapse. The gold-standard way of obtaining ground-truth measurements of activity intensity is using video as a proxy for direct observation\citep{keadle2019framework}. Historically, battery limitations have meant that there has been a trade-off between the temporal resolution, and total duration recorded. For instance, \citep{keadle2024using} used GoPros to record two sessions of two hours of free-living data in a study of 26 participants. On the other hand, the studies considered in this work have recordings covering 8+ hours in over 100 participants each, though at the expense of only capturing images every 20+ seconds. In Table \ref{tab:validation_studies}, we highlight the sizes of comparable camera based validation studies, and there is a notable gap between the size of studies achieved using video compared to time-lapse recordings.

\subsection{CAPTURE24: the Oxfordshire and Sichuan studies}
\label{sec:capture24_studies_review}
The CAPTURE24 study was collected in 2014 from 165 participants in the Oxfordshire county of the United Kingdom in order to validate wrist-worn accelerometer-based physical activity measurement approaches in adults \citep{gershuny2020testing, chan2024capture}.  The CAPTURE24-CN study was collected in 2017 from 113 participants in the Sichuan province of China alongside a similar effort to develop and validate approaches to derive wrist-worn accelerometer-based physical activity measurements in over 20 000 participants in the China Kadoorie Biobank\citep{chen2023device}. Though these studies only comprise roughly 100 participants each, they are the primary source of labelled data used to validate the measurements conducted in large scale health studies such as the UK and China Kadoorie Biobank \citep{doherty2017large, willetts2018statistical, doherty2018gwas, walmsley2022reallocation}, comprising tens of thousands of participants. 
As highlighted in Table \ref{tab:validation_studies}, they represent the largest available validation studies.

\subsection{Recognising activities from sparse sequences of egocentric images}

Collecting and analysing data using wearable cameras has a history spanning over three decades, with pioneering work by Mann \citep{mann1997wearable} and Aizawa \citep{aizawa2001summarizing}, but was also foreseen as early as 1945 \citep{bush1945we}. There have been several works which explore human activity recognition in third person \citep{feichtenhofer2019slowfast, zhang2022actionformer, momeni2023verbs, keadle2024using}, and, to a lesser extent, egocentric videos \citep{grauman2022ego4d, lin2022egocentric, pramanick2023egovlpv2}. Working towards the goal of reducing annotation burden in wearable data-sets, \cite{bock_weak-annotation_2024} proposed a clustering-based strategy where annotators label a representative clip in clusters of similar clips, derived from vision-foundation model features \citep{radford2021learning, oquab_dinov2_2024, carreira_quo_2017}, which is then applied to all clips within each cluster. In contrast, we focus on methods which do not require human input, and which work on sparse sequences of images. 

There has been some prior work on human activity recognition from sparse, egocentric sequences of images \citep{wang2013using, moghimi2014analyzing, castro2015predicting, cartas2017recognizing, cartas2020activities, cartas2021understanding}, though in datasets with only 10s of participants. These works focus on training discriminative models to predict predefined sets of labels, but the variation in how these labels are defined, and lack of publicly available benchmarks, makes it difficult to compare results across different works. 

Though there has been less work on modelling activity from sparse sequences of egocentric images seems over the past few years, there has been increased interest in modelling egocentric video, spurred on by a number of relatively large, open-source data-sets, such as EPIC-KITCHENS \citep{damen2022rescaling}, Ego4D \citep{grauman2022ego4d}, and Ego-Exo4D \citep{grauman2024ego} which move away from being labelled by sets of predefined activities towards open-ended natural language descriptions.

\subsection{Vision language models}

Vision-Language Models (VLMs) are a broad class of models which process both visual, and textual data for tasks such as image-based text retrieval, image captioning, and image classification \citep{li2024multimodal}. Natural language descriptions of visual content, such as alternative text descriptions of images, or summaries of video segments, are widely available on the internet, sidestepping the need for annotated data. VLMs, such as CLIP \citep{radford2021learning}, and LLaVA \citep{liu2024llava}, are typically trained on large data-sets of pairs of images and text, scraped from the internet, such as WebImageText \citep{radford2021learning} and LAION-5B \citep{schuhmann2022laionb}, and increasingly, synthetic labels generated by frontier multimodal models, such as GPT-4, are used to make up higher quality data-sets in a secondary training stage \citep{liu2024llava}. Despite having not been explicitly trained for them, these models have shown good performance in several downstream tasks, including image classification on benchmarks such as ImageNet \citep{deng2009imagenet}, suggesting that pretraining VLMs on large data-sets produces models which transfer well to new tasks. One recent work suggests the success of VLMs in recognising concepts in downstream tasks can be attributed to the prevalence of these concepts in their large pretraining data-sets, though with the performance scaling logarithmically with concept frequency \citep{udandarao2024no}. 

In this work, we consider both a dual encoder VLM, CLIP \citep{radford2021learning}, which quantifies the similarity between images and text, and generative VLMs, BLIP2 and LLaVA, which can be prompted to describe, and answer questions about images. All of these models have mechanisms which allow them to perform image classification in a ``zero-shot'' transfer setting, i.e. without having seen task-specific data, in this case, egocentric images labelled with activity intensity classes.

\section{Methods}

Our aim was to assess the performance of VLMs for predicting physical activity behaviours from wearable camera images. To do this, we compared the performance of different VLMs and discriminative models on two free-living validation studies labelled with labels from the compendium of physical activity, which have known mappings to activity intensity classes. 

\subsection{Data processing and quality assessment}
\label{sec:data_process}

The Oxfordshire and Sichuan validation studies were primarily developed to validate accelerometer based measurement of physical activity. Thus, there has not been a detailed exploration of the wearable camera portion of the data-set, vital as this is for informing the labels that train and test accelerometer-based approaches. The images in these data-sets are egocentric, which means there is inherent ambiguity in the participant's activities, since the participants themselves remain largely unobserved. In addition to ambiguity introduced by the camera perspective, there is ambiguity introduced by the low, variable frame rate and by the camera being occasionally obstructed, or taken off. All of these factors influence how well annotators were able to label the data. In Appendix \ref{sec:properties_timelapse}, we explore the relationship between image capture rate and the number of activities that could be distinguished for each participant, and image obscurity, related to the darkness and variation of each image, against whether the image was annotated.

Images in both studies that were not labelled were excluded from the rest of our analysis, and we indicate the number of labelled images in each study in Table \ref{tab:demographics}. Based on the large number of unannotated images in the Sichuan data-set, we decided not to do model development on this data-set, and purely reserve it for model testing. $70\%$ of the participants in the Oxfordshire study were randomly selected for model training, $15\%$ for validation and model selection, and $15\%$ for testing the final models.

\subsubsection{Simplifying labels}
Both validation studies were annotated using a modified version of the 2011 compendium of physical activity \citep{ainsworth2011compendium}, which organises labels in a hierarchy such as, ``transportation;walking;12150 running'', each associated with a metabolic equivalent of task (MET) value, which estimates the ratio of the activity's metabolic rate to a standard resting metabolic rate of $1\text{kcal}\cdot kg^{-1}\cdot h^{-1}$ \citep{ainsworth2011compendium}. Instead of using the exact MET values, we mapped each activity to one of three activity intensity classes, defined as:
\begin{description}
    \item[Sedentary Behaviour] (SB) waking behaviour at $\leq$ 1.5 METs in a sitting, lying or reclining posture,
    \item[Light intensity physical activity] (LIPA) waking behaviour at <3 METs not meeting the sedentary behaviour definition,
    \item[Moderate-vigorous physical activity] (MVPA) waking behaviour at $\geq 3$ METs, and
    \item[Sleep] Non-waking behaviour (not used in this work, though included for completeness).
\end{description}
These definitions are in line with the definition of SB obtained through consensus in \cite{tremblay2017sedentary}, and the definitions of LIPA and MVPA used by \citep{ainsworth2011compendium, keadle2024using}. We report the median of the number of images in each intensity class per participant in Table \ref{tab:demographics}, and show the spread in the per-participant tallies as quartile plots in Figure \ref{fig:label_imbalance}. 

When doing exploratory data analysis, we noticed that some of the raw labels were misspelled, e.g. ``office wok/computer work general'', and that the same activities would be included in multiple labels with different prefixes, such as ``walking;5060 shopping miscellaneous, and ``5060 shopping miscellaneous''. To come up with a more concise set of labels, we used a sentence embedding model \citep{reimers2019sentencebert} to embed the labels, and then used agglomerative clustering to build a dendrogram of related labels, based on their embeddings \citep{hastie2009elements, scikit-learn}. We then manually went through the tree, merging sets of labels with the same meaning together. We refer to this concise, semantically deduplicated set of labels as the `clean labels'. This set of labels represents a more detailed set of colloquial activities encompassing the activities performed in the Oxfordshire study, which we use in Section \ref{sec:generative_activity_prediction} as an intermediate set of targets when predicting activity intensity.

\subsection{Predicting activity intensity using computer vision}
\label{sec:model_details}
In order to asses how well computer vision methods can predict activity intensity classes from wearable cameras, we went through a process of model training, hyperparameter tuning, model selection and testing on data from unseen participants. We considered two different discriminative and three different VLMs, and for each model, we conducted a random search over the model hyperparameters \citep{goodfellow2016deeplearning}, evaluating the performance of each hyperparameter run on the validation split. Finally, we selected the best discriminative model, and VLM, and evaluated their performance on the test split of the Oxfordshire study, and on the entire Sichuan study.

Given an image as input, the discriminative models output a vector, indicating the probability of the image belonging to one of the 3 activity intensity classes. The VLMs can further be divided into generative models, which output natural language descriptions given an image and an optional prompt as inputs, and dual-encoder models, which embed each image and a natural language description of each class into a joint embedding space, where the similarity between different images and descriptions can be quantified by looking at the similarity between their embeddings.

We investigated two generative VLMs, 3 billion parameter BLIP2 \citep{li2023blip}, based on the FlanT5-XL language model \citep{chung2022scaling}, and 7 billion parameter LLaVA \citep{liu2024llava}, and one dual-encoder model, CLIP \citep{radford2021learning}. We used the model checkpoints available on Hugging Face \citep{wolf2019huggingface}, and the exact Hugging Face model IDs are given in Table \ref{tab:model_implement}. BLIP2 and LLaVA are both open-source VLMs which have shown strong performance on image captioning, with both adopting the CLIP vision encoder as a component, motivating the inclusion of CLIP as a stand-alone model to ablate the benefits of using prompted, generative VLMs, which include language models as an additional component, over a dual-encoder model. 

We tested these VLMs against a commonly adapted transfer learning approach of fine-tuning a pretrained model using task specific data, and we refer to the resulting models as discriminative models. As a baseline model, we used a ResNet-50 \citep{he2016deep}, pretrained on ImageNet-1K \citep{deng2009imagenet}, and the image encoder from CLIP, pretrained on WebImageText \citep{radford2021learning}, which we refer to as ViT, which is a reference to its vision transformer architecture \citep{dosovitskiy_image_2021}. Though the focus of this paper is on image based classification, we also include the best sequence model found in \cite{cartas2020activities}, ResNet-LSTM, which has the advantage of being able to access information from multiple images. 

\subsubsection{Discriminative models}
For the discriminative models, we trained the models on the training split, monitoring performance on the validation split throughout training. We used the AdamW optimizer \citep{loshchilov2018decoupled} to update model weights to minimise the cross-entropy loss, and used early stopping to terminate the training, monitoring the validation cross-entropy loss, with a patience of 5. The best model found during training based on the validation loss was used to made predictions on the validation split, from which we calculated the validation metrics used to perform model selection, and study the impact of hyperparameters. For all models, we replaced the final fully connected layer of the image encoders. For the single image models, ResNet and CLIP image encoder, we replaced it with a linear layer with three outputs. The ResNet-LSTM was constructed by using a Long Short-Term Memory unit \citep{hochreiter_long_1997} to model temporal dependencies across 3 independently encoded image embeddings produced by a ResNet-50 \citep{he2016deep}.

One of the most important hyperparameters for discriminative models is the learning rate \citep{goodfellow2016deeplearning}, and for all the single-image based discriminative models we did a random search over different learning rates, batch-sizes, whether we applied data-augmentation, and whether we did full fine-tuning, or only fine-tuned the linear layer. For each model, we did 30 trials of different hyperparameters. The search space for these hyperparameters is presented in Table \ref{tab:model_hparams}, and the exact sweep configurations for each model are in the repository. The only hyperparameter tuning done for the ResNet-LSTM was to train three different models with learning rates, $10^{-3}, 10^{-4}, 10^{-5}$. For data-augmentation, we used TrivialAugment, which samples a single augmentation uniformly at random from a set of 21 augmentations, along with a strength with which the augmentation is applied to each image \citep{muller_trivialaugment_2021}

\subsubsection{Dual-encoder CLIP}
As proposed in \cite{radford2021learning}, we classify images by embedding them using the image encoder, and the set of labels using the text encoder. Classification is then framed as a text retrieval task where for each image, we retrieve the most similar label by looking at the cosine similarities between each image embedding, and all the label embeddings, and selecting the label associated with the largest cosine similarity.

We either used natural language descriptions of the intensity classes as targets, or used the more detailed clean labels as targets, which have a known mapping to the intensity classes. Intuitively, the set of clean labels represent more colloquial descriptions of physical activity, which may be better represented in the pretraining data-sets of VLMs compared to the intensity classes. For instance, the phrase
``sedentary behaviour'' might not be well represented, whereas phrases such as ``lying down'' which represent instances of SB, might be more prevalent. When using the intensity classes as targets, SB was represented as ``sedentary behavior'', LIPA as ``light physical activity'', and MVPA as ``moderate-to-vigorous physical activity''.

A similar idea of adapting pretrained VLMs by rephrasing the text targets was explored in \cite{mirza2023lafter}, where they used a large language model to generate alternate descriptions for each of the target labels and trained a linear classifier to map between embeddings of the target labels and embeddings of the corresponding alternate descriptions. Our approach can be viewed as a non-parametric alternative to this. However, a weakness with both of these approaches is that neither of them strictly check whether an intensity class is implied by the generated description, and we show some of these failure cases in Table \ref{tab:qualitative_ex}.

\subsubsection{Generative models}
\label{sec:generative_activity_prediction}

For the generative VLMs, we used different prompts to condition text generation. To evaluate whether the true intensity class could be inferred from the model's natural language description of each image, we used a text-embedding model, \href{https://huggingface.co/sentence-transformers/all-MiniLM-L12-v2}{all-MiniLM-L12-v2}, to embed the descriptions \citep{reimers2019sentencebert}, and then followed a similar strategy to CLIP of mapping these descriptions to either the nearest intensity class, or the nearest clean label based on the similarity of their embeddings. In addition to varying the mapping approach, we varied the number of tokens generated, the prompt, and how we represented the activity intensity classes. We proposed an initial set of prompts, ranging from task-specific ones, e.g. ``Question: What is the intensity of the physical activity in the image? Options: Sedentary, Light, Moderate-Vigorous. Short answer:'', to more generic descriptive prompts, e.g. ``a photo of''. We also augmented the set of prompts by asking proprietary large language models, ChatGPT, Claude, and Gemini, to suggest similar prompts and selecting sensible ones. The final set of 17 prompts is included in the repository. The exact hyperparameters that were varied for each model are shown in Table \ref{tab:model_hparams}.

\subsection{Evaluation}
We assessed each model's performance across activity intensity classes using Cohen's $\kappa$ score, and the performance per class using the F$_1$-score of the class \citep{scikit-learn}. The Cohen's $\kappa$ score ($\kappa$ or ``kappa'' for short in Figures) is 0 if the model's performance is on par with a random classifier, and 1 if all instances were correctly predicted. The F$_1$-score for a class is the harmonic mean of the recall, the proportion of instances of the class that were correctly predicted, and the precision, the proportion of predictions of that class that were correct. Since there is a class imbalance, reporting per class F1-scores helps avoid inflating the performance of classifiers that are biased towards predicting the majority class.  We calculated these metrics per participant and present the spread of the per-participant scores in our results. This does however come with the caveat that some participants had relatively few instances of LIPA and MVPA, thus the estimate of these metrics at the participant level had high variance. 

\section{Results}
In Section \ref{sec:data_results}, we present the results from data-processing and exploratory data analysis, highlighting some of the challenges of modelling free-living egocentric timelapses, and in Section \ref{sec:model_results}, we present results from model selection, motivating the choice of the best models. Finally, we present the performance of the best vision-language and discriminative model.

\subsection{Data processing and EDA}
\label{sec:data_results}
\begin{table}
    \centering
    \caption{Summary statistics for each data-set, comparing the size, resolution and demographics between the Oxfordshire and Sichuan study}
    \begin{tabular}{rcc}
        \hline
        ~ & Oxfordshire&Sichuan\\
        \hline
        Number of participants &  161 & 111 \\
        Number of labelled images (\% all images)& 231 837 (74\%)&46 184 (34\%)\\
        Median $\delta t$ (1st, 3rd quartile) between images (s)& 24 (23, 32)&84 (69, 88)\\
        No. unique labels& 220&110\\
        \hline
        Median instances per participant: Sedentary & 884 & 184 \\
        LIPA & 441.5 & 142 \\
        MVPA & 81 & 45 \\
        \hline
        Number of participants (\%) aged: 0-30 & 45 (28\%)& 12 (11\%)\\ 
        30-50 & 67 (42\%)& 43 (41\%)\\ 
        50-70 & 39 (24\%)& 49 (47\%)\\ 
        70-100 & 8 (6\%)& 1 (1\%)\\
        \hline
        Sex: Female & 103 (64\%)& 63 (58\%)\\ 
        Male & 58 (36\%)& 45 (42\%)\\
    \hline
    \end{tabular}
    
    \vspace{0.5em}
    \footnotesize
    \textit{Note:} There were no reported ages for 2 participants in the Oxfordshire study. In the Sichuan study, 4 participants had no reported age, 2 had invalid ages ($\geq$500), and 3 had no reported sex.
    \label{tab:demographics}
\end{table}

The Oxfordshire study had 231 837 (from an original 312 585) images with non-trivial labels from 161 participants (Table \ref{tab:demographics}), i.e not labelled as "uncodeable", or "undefined".  The median time interval, $\delta t$, (1st, 3rd quartile) between images was 24 seconds (23, 32). The Sichuan study had a much larger median time interval of 84 seconds (69, 88), and a much smaller proportion of images with non-trivial labels of 46 184 images (from an original 132 850 images) from 111 participants. 

We estimated the time covered in each study as
\begin{equation*}
\text{time covered (h)} = \frac{
\text{no. labelled images} \times \text{median}\, \delta t \, \text{between images (s)}
}
{
60\times60
},
\end{equation*}
suggesting that there were 1 546 hours of labelled data in the Oxfordshire study and 1 078 hours of labelled data in the Sichuan study, though this is an overestimate because the low temporal resolution, particularly in the Sichuan study, means that knowing the activity in each image does not necessarily mean we continue to know the activity in an 84-second window surrounding that image.

One noticeable feature of both data-sets is the large number of images that were difficult to label. We differentiate between images that were unlabelled, and images where the labels were unknown, which includes both unlabelled images, and images with labels such as "image dark/blurred/obscured". Although the number of unlabelled images in both study was relatively low (7.57\% for the Oxfordshire study and 1.31\% for the Sichuan study), the number of images with unknown labels was very high (25.8\% for the Oxfordshire study and 65.2\% for the Sichuan study).

The median $\delta t$ between frames was much lower in the Sichuan study, compared to the Oxfordshire study. In the appendix, Figure \ref{fig:events_time}, echoes this, though by showing the median $\delta t$ for each participant, also reveals that participants clustered around four distinct median capture rates, suggesting that different base capture rates were erroneously set on the Autographers, leading to these different resolutions. Although the estimated number of hours captured in each study are of similar orders of magnitude, the number of annotated events in the Sichuan study is much lower, pointing to the lower capture rate set on the devices as being a bottleneck for the resolution of the annotations.

\subsection{Model results}
\label{sec:model_results}

\begin{figure}[ht]
    \centering

    \begin{subfigure}[b]{\textwidth}
    \centering
    \includegraphics{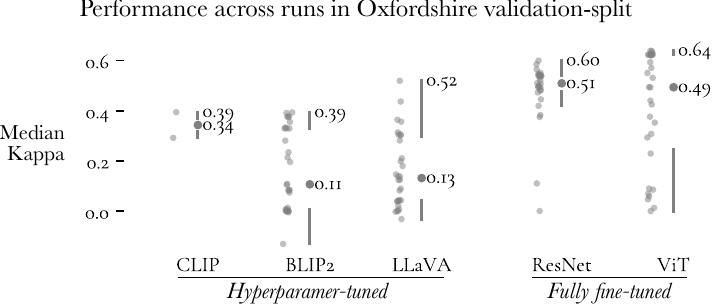}
    \caption{Quartile plots show the range of median $\kappa$s on the Oxfordshire validation-split across the 30 runs for each model (except for CLIP). Each run randomly sampled a different set of hyperparameters. The median $\kappa$ of each run is shown as a jittered column of dots to the left of each quartile plot. The maximum of the median $\kappa$s is indicted above the quartile plot, indicating the performance of the best found hyperparameters for each model, and the median is indicated to the right.}
    \label{fig:val_runs}
    \end{subfigure}

    \begin{subfigure}[b]{0.48\textwidth}
    \centering
    \includegraphics{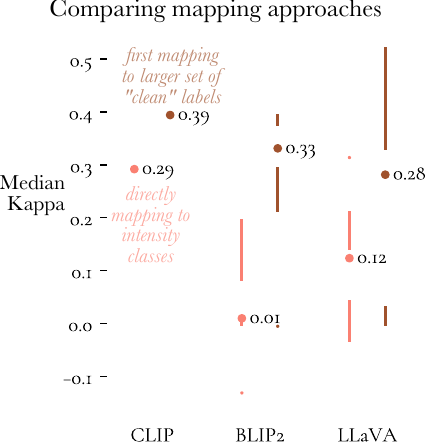}
    \caption{Quartile plots comparing directly calculating the similarity between the generative model image descriptions, or image embeddings for CLIP, and the intensity labels, versus calculating their similarity to a broader set of activities with known mappings to the activity labels. The results reflect the spread of the median $\kappa$ across runs with different randomly sampled prompts, and number of tokens generated.}
    \label{fig:mapping_approach}
    \end{subfigure}
    \hfill
    \begin{subfigure}[b]{0.48\textwidth}
        \centering
        \includegraphics[width=\textwidth]{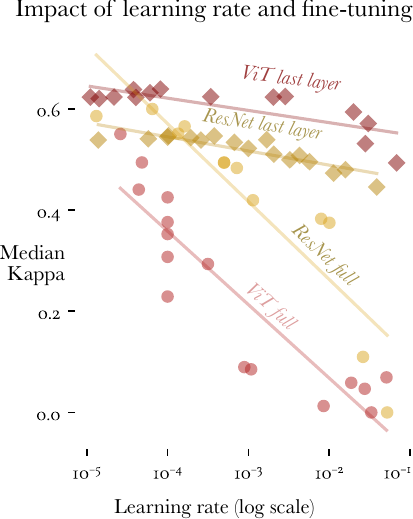}
        \caption{Scatter plot showing the median $\kappa$ of runs with different learning rates, and where either only the last layer, or the full model was fine-tuned. Median $\kappa$s were higher for ViT than ResNet runs when only the last layer was fine-tuned, and considerably worse when fine-tuning the full model.}
        \label{fig:sublte_finetuning}
    \end{subfigure}
    \caption{Impact of different hyperparameters on the performance of each model on the validation-set of the Oxfordshire study.}
    \label{fig:hparam_insights}
\end{figure}

We used the model's validation performance on the Oxfordshire study to identify promising models, and for each model, promising hyperparameters. The left side of Figure \ref{fig:val_runs} shows that for the VLMs, differences in the prompts, mapping approach, and number of generated tokens resulted in large differences in validation performance ($\kappa$ scores range from 0 to 0.5). The right side of Figure \ref{fig:val_runs} shows the validation performance of fine-tuned DMs, which tended to be better than the VLMs, though also displays a sensitivity to different hyperparameters. 

For the VLMs, we highlight the mapping approach as one of the hyperparameters associated with this variation. Figure \ref{fig:mapping_approach} visualises the difference in performance between runs that used the larger-set of more colloquial activities as targets and those which directly used SB, LIPA, and MVPA as targets. Across all VLMs, the median performance of the runs that adopted the more colloquial targets was higher. Despite this, the best performing VLM, LLaVA, which was prompted, ``Walking, Running, Sitting, Standing, Other. Based on the objects in the image, what is the person likely doing?'', had its responses directly mapped to one of the activity classes, and not the clean labels.

Examining the spread in validation performance across different hyperparameter runs for the ResNet and ViT in isolation suggests that the ResNet is the more robust model, since the median of the median $\kappa$ scores is higher, and the interquartile range is narrower. Figure \ref{fig:sublte_finetuning} elaborates on this picture, revealing that the combination of doing full fine-tuning and using a high learning rate ($l \geq 10^{-4}$) was particularly detrimental for the ViT, and that when when only fine-tuning the last layer, the performance of the ViT was consistently better than the performance of the ResNet. We saw better performance from fine-tuning the last layer as opposed to full fine-tuning, despite the latter being a more flexible model adaptation technique. In general, lower learning rates were associated with better validation performance, with the relationship between the logarithm of the learning rate and the median $\kappa$ roughly following a negative linear line, suggesting that performance could be further improved by using even lower learning rates.

Finally, we selected the best performing vision-language (LLaVA), and discriminative model (ViT), and assessed their performance on the withheld test-set (Figure \ref{fig:main}). SB in the Oxfordshire test-set was well predicted by all models, with median F$_1$-scores of 0.89 (0.84, 0.92) for LLaVA and 0.91 (0.86, 0.95) for ViT. Predictive performance on LIPA and MVPA, although much better than chance performance, was worse than SB, which a median F$_1$-score of 0.60 (0.56,0.67) for LLaVA, and 0.70 (0.63, 0.79)) and for ViT. The spread in the performance across participants was large for these behaviours, particularly MVPA. We found a large drop in performance when going from the Oxfordshire study, where models were trained and/or hyperparameter-tuned, to the Sichuan study. The largest drop in performance was for the ViT, which went from a median $\kappa$ of 0.67 (0.60, 0.74), which can be interpreted as showing substantial agreement relative to the human annotations \cite{landis_measurement_1977}, to 0.19 (0.10, 0.30), which only shows fair agreement. For LLaVA the drop in performance was from a median $\kappa$ of 0.54 (0.64, 0.49) to 0.26 (0.15, 0.37).

Whereas human annotators were allowed to view the entire history of a participant's day when annotating each image, these models make predictions based on single images. In order to estimate human performance in the same setting, one of the present authors manually labelled $>500$ randomly selected images from the test-set of each study, without temporal context, and obtained a median $\kappa$ of 0.63 (0.45, 0.72) on the Oxfordshire study, and 0.572 (0.46-0.61) on the Sichuan study. The performance on the Oxfordshire study is similar to the performance observed for the best model, though noticeably better than the model performance of the Sichuan study.

Though not strictly a fair comparison to the single-image models, we also tested the performance of a sequential model (ResNet-LSTM) to investigate the benefits of going beyond single frame predictions. This model consistently had similar or slightly better F$_1$-scores for each of the activity intensity classes compared to the best single-image model, and obtained a median $\kappa$ of 0.66 (0.59, 0.72) on the Oxfordshire study and median $\kappa$ of 0.31 (0.18, 0.41) on the Sichuan study suggesting that performance can be further improved by developing sequential models for these sparse sequences of images.

 Finally, if we look at the accuracy of the models, which is misleading in that it is dominated by performance on the majority class, but relevant in that it relates to the fraction of images that would have to be corrected by a human annotator, both of the best models achieve an accuracy $> 80\%$ on the Oxfordshire test-set, and $> 50\%$ on the Sichuan study.

\section{Discussion}
We compared the performance of VLMs and DMs on predicting activity intensity in two free-living validation studies, and found that SB was well predicted in unseen participants within the Oxfordshire study, but that LIPA and MVPA were less well predicted, and all models generalised poorly to the Sichuan study. The overall accuracy of the models on unseen participants in the Oxfordshire study suggest they might still be useful for labelling wearable camera images, especially in free-living data where SB typically makes up the majority of instances as seen in Table \ref{tab:demographics}, though within similar studies to ones they have been adapted for.

Similar work by \cite{keadle2023evaluation}, though based on third-person still frames from a GoPro, found that their best model at distinguishing between SB, light, moderate, and vigorous intensity physical activity, a tree-based model (XGBoost,  \cite{chen2016xgboost}) based on features from AlphaPose \citep{fang2022alphapose}, was able to do so with an accuracy of 68.6\%. Although they separate out moderate and vigorous physical activity into distinct classes, we can calculate performance metrics compatible with this work by combining the rows and columns for these classes in the confusion matrix in Table 3 of their work, included here in Table \ref{tab:keadle_conf_mat}, comparing it to the confusion matrix in Figure \ref{fig:conf_mat}. 

The overall accuracy for predicting activity intensity of XGBoost was 69.2\%, compared to the finetuned ViT in this work, which achieved an accuracy of 84.6\% on unseen data in the Oxfordshire study, and LLaVA, which achieved an accuracy of 80.9\%. The improved performance of ViT and LLaVA in this context is in part driven by better recall of SB, which was predicted with a recall of 71.6\% in \cite{keadle2024using}, but with recalls of 90.7\% and 89.1\% for ViT and LLaVA, respectively, in this work, and there was also a higher proportion of SB in studies used in this work, thus the accuracy was more heavily weighted by SB. If we consider the average of the per-class recalls, which weights the classes equally, the performance is closer, 70.0\% for XGBoost, 76.8\% for ViT and 72.5\% for LLaVA.

However, there are many limitations to this comparison, including the varying perspectives (first vs. third person), and frame-rates (0.05 vs. 30 fps) with which each study captured footage. Annotating activity intensity classes from third person video recordings is considered the gold-standard for validating device-measured activity intensity measurements \citep{keadle2019framework}. \cite{martinez_validation_2021} compared using sparse sequences of images captured by wearable cameras to assess posture against the activPAL and reported that, although the bias in estimates of sitting time was not significant, there was significant bias in estimates of standing and movement time. Figure \ref{fig:2dhist} demonstrates the difficulty in interpreting images in this regime, with a large number of dark images with  low variance remaining unannotated, a common limitation of this type of data capture. On the other hand, the use of egocentric cameras for capturing validation data is more scalable since it does not require researchers to follow participants, enabling the Oxfordshire and Sichuan validation studies to collect data from 100+ participants each.

The focus on models based on single images was motivated by the availability of VLMs in this setting, and the lack of models for sparse sequences of images. However, predicting activities from single images is a notable obstacle, and our limited analysis of one annotator's performance in this regime suggests that the current levels of performance on the Oxfordshire study are close to human performance based on single images. Beyond single-image models, the ResNet-LSTM, performed slightly better than the single-image models, and did not undergo hyperparameter tuning to the same extent. This suggests the necessity of moving beyond single-frame models. This was an imbalanced problem, and we observed high variation in the performance estimates of the less prevalent classes. Our performance estimates could have been more robust by adopting methods such as cross-validation, though at the expense of these experiments being more computationally expensive. Each hyperparameter-tuning run took an average of 5 hours to complete on a V100 GPU for the ResNet, the smallest model. 

Despite these limitations, this work was able to assess performance in studies collected in free-living conditions in a large number of participants revelative to existing wearable validation studies, and it assessed generalisation using an independently collected study. Activity intensity classes have been adopted in a number of downstream epidemiological works \citep{walmsley2022reallocation, schalkamp_wearable_2023, shreves_dose-response_2023}, and we used definitions compatible with this field of research. The application of VLMs to estimating activity intensity is novel, and also raises the possibility of measuring new behaviours, such as environmental exposures, social interactions, eating and drinking behaviours, without the need for task specific training. An application using VLMs to label outdoor time to validate wrist-worn light sensors is concurrently being explored. 

Improvements in technology not only suggest new ways of analysing validation studies, but also conducting them. \cite{tran2024memorilens} proposed developing wearable cameras which cost less, and \cite{mamish2024nir} proposed a wearable camera able to capture footage at high frame-rates while lasting several days. Commercially available body cameras, such as those manufactured by BOBLOV and MIUFLY, are commercially available and able to record ~15 hours of video footage on a single charge. The adoption of these cameras in future validation studies would reduce the annotation uncertainty due to low frame-rates whilst making it easier to adopt activity recognition approaches developed for egocentric video \cite{pei_egovideo_2024}. Although we focus on wearable cameras as a way of informing ground truth labels to validate and train measurement approaches typically using other wearable sensors, wearable cameras have also been used in small health studies \citep{doherty2012use, kerr2013using, gage2023fun} as the measurement device themselves. Given the range of behaviours that can be measured simultaneously from a single camera in comparison to other wearables, and the human interpretable nature of the modality, one might be tempted to directly adopt them in health studies. However, the large amount of information captured by these cameras raises various ethical issues, and has made it unlikely that they will be adopted for large scale health studies \citep{mok2015too, meyer2022using, kelly2013ethical}. 

Although we have taken the distinction between the broader field of activity recognition and recognising health relevant activity intensity classes, progress in the former is vital to this task, and should not be disregarded. This work showed that the performance of generalist VLMs is similar to domain specific discriminative models, and progress on developing more capable generalist models might well outpace approaches reliant on annotated wearable data. This suggests the importance of exploring similarities between more mainstream computer vision research and the present study. There is also additional work needed in applying methods from fields such as continual learning, active learning and uncertainty quantification so that models can be adapted and assessed `on the fly' to efficiently learn from new labelled data, so that human input can be used efficiently in correcting the most informative instances, and so that models can indicate which samples they cannot reliably label. After all, model accuracy is only one aspect impacting the efficiency of labelling wearable data-sets.

\section{Conclusions}
In this paper we assessed the performance of fine-tuned discriminative models and vision-language models on the simple, but important task of predicting activity intensity classes from two free-living validation studies, each comprising over 100 participants, conducted in Oxfordshire, UK, and Sichuan, China. Sedentary behaviour was well predicted within unseen participants from a seen population by both types of models. Random searches over different hyperparameters revealed the importance of how activity intensity classes were phrased when using vision-language models, and the importance of minimal fine-tuning for the discriminative models.  Although none of these approaches pass the threshold required for trained human annotators, we only focused on activity prediction based on single images, which is a notable handicap on model performance, and initial results reproducing a sequence-based classifier in this setting shows slightly better performance. Although several times bigger than existing validation studies, the studies used here were still prone to errors in the ground-truth labels arising from the sparsity of the images, and large numbers of obscure images. Despite these limitations, we would recommend the adoption of the best models found in this study to label sedentary behaviour in free-living studies as they are freely available, relatively easy to adapt and can substantially reduce the annotation burden given the prevalence of sedentary behaviour. We would also encourage research groups conducting wearable camera based validation studies to consider moving to newer wearable cameras which are able to record videos for the full waking day, which would significantly lower the uncertainty in the ground-truth labels of physical activity.

\section*{Funding}
Abram Schönfeldt is supported by the EPSRC Centre for Doctoral Training in Health Data Science (EP/S02428X/1). Aiden Doherty’s research team is supported by a range of grants from the Wellcome Trust [223100/Z/21/Z, 227093/Z/23/Z], Novo Nordisk, Swiss Re, Boehringer Ingelheim, National Institutes of Health’s Oxford Cambridge Scholars Program, EPSRC Centre for Doctoral Training in Health Data Science (EP/S02428X/1),  British Heart Foundation Centre of Research Excellence (grant number RE/18/3/34214), and funding administered by the Danish National Research Foundation in support of the Pioneer Centre for SMARTbiomed. Xiaofang Chen acknowledges support from the Noncommunicable Chronic Diseases–National Science and Technology Major Project (2023ZD0510100) and the National Natural Science Foundation of China (82192900, 82192901, 82192904, 81390540, 91846303). For the purpose of open access, the author(s) has applied a Creative Commons Attribution (CC BY) licence to any Author Accepted Manuscript version arising.

\section*{Acknowledgements}
Thank you to Shing Chang, Hang Yuan, Aidan Acquah, Laura Brocklebank, Jerred Chen and Freddie Bickford Smith for valuable advice over the course of this project. We are grateful to Huaidong Du for facilitating access to the Sichuan validation study, and we extend our thanks to all those involved in the collection and annotation of the validation datasets. Finally, we are  grateful to the participants for their willingness to participate in these studies.

\section*{Author contributions}
Abram Schönfeldt led the study design, data analysis, and drafting of the manuscript, and contributed to project conceptualization. Ronald Clark contributed to project conceptualization, provided supervision, and reviewed and suggested edits to the manuscript. Ben Maylor provided supervision, offered technical guidance, and reviewed and suggested edits to the manuscript. Xiaofang Chen contributed to data collection. Aiden Doherty contributed to project conceptualization, supervised the study, and reviewed and suggested edits to the manuscript. All authors read and approved the final version.

\section*{Conflicts of interest}
The authors declare no conflicts of interest.

\appendix

\section{Appendix}
Section \ref{sec:mapping_details} includes additional details when mapping the labels to activity intensity classes. Section \ref{sec:properties_timelapse} goes into more detail on the properties of the validation studies used in this work, and Section \ref{app:implementation} provides additional implementation details. Section \ref{app:more_results} shows confusion matrices of the best models, and illustrates examples of generated captions mapped to different activity classes, and Section \ref{app:human_performance} presents median $\kappa$ scores of one annotator confined to predicting activity intensity from single images on a subset of the data.

\subsection{Mapping compendium annotations to activity intensity classes}
\label{sec:mapping_details}

This mapping from the applied compendium of physical activity labels to activity intensity classes  was originally done in \citep{walmsley2022reallocation}. Note, however, that the published dictionary does not strictly abide by these definitions, since some activities which technically would be MVPA, such as ``Cleaning, sweeping carpet or floors, general'', MET = 3.3, were mapped to LIPA based on the discretion of the authors. To be consistent with previous work using the Oxfordshire study, we used this mapping, also applying it to the labels in the Sichuan study accounted for by it.

There were some labels used in the Sichuan study not included in the dictionary from \cite{chan2024capture}. To address these, an updated dictionary was created using the 2024 compendium of physical activity \citep{herrmann2024Compendium} by matching the raw labels to their updated entries using their activity codes. This dictionary provides an updated mapping from the raw labels from both validation studies to activity intensity classes, and the latest entries in the compendium of physical activity and will be made available with the supplementary material.

\subsection{Properties of two free-living, egocentric timelapse}
\label{sec:properties_timelapse}

\begin{figure}
    \centering
    \includegraphics[width=\linewidth]{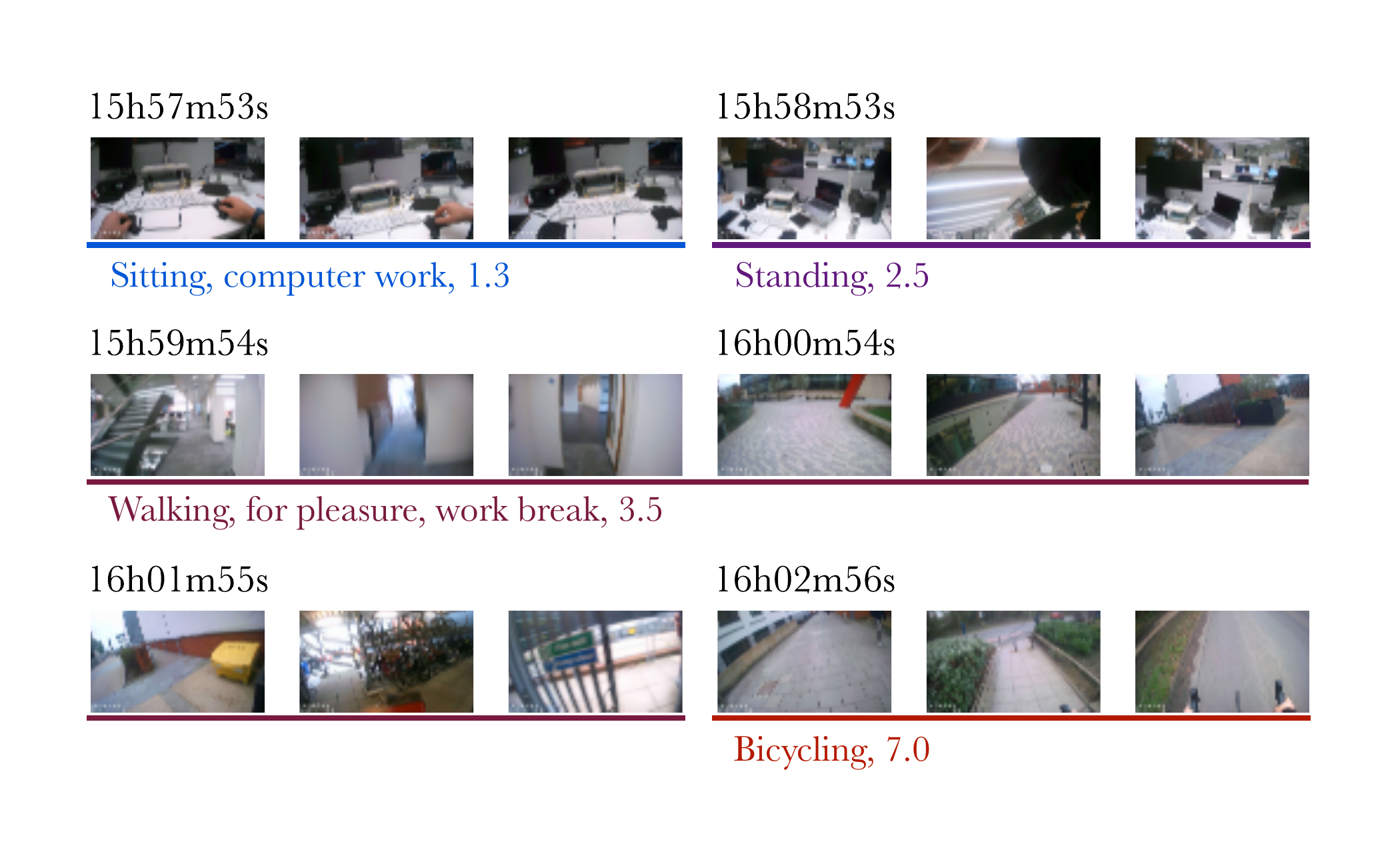}
    \caption{A sequence of images captured with an interval of 20 seconds between frames, labelled with activities and MET values.}
    \label{fig:camera_20s}
\end{figure}

The two studies used in this work used wearable cameras capturing sparse sequences of images to label activities of daily living. Figure \ref{fig:camera_20s} provides an example of a sequence of activities captured by a wearable camera with a time interval of 20 seconds between consecutive frames. At this frame rate, the transition between environments can be abrupt, and the segment of cycling only becomes apparent once the handlebars are visible a few frames after the start of the event.

\begin{figure}
    \centering
        \begin{subfigure}[b]{0.48\textwidth}
        \centering
        \includegraphics[width=\textwidth]{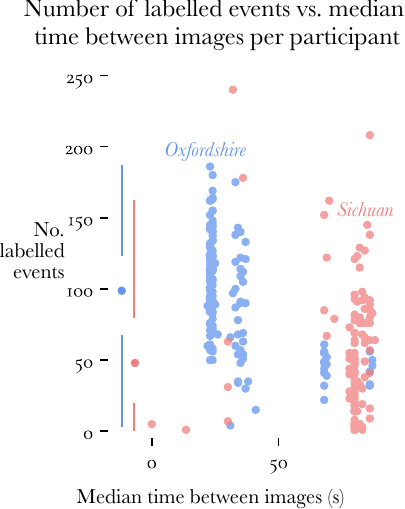} 
        \caption{Scatter plot showing that most participants within the Oxfordshire study had a lower median time between images compared to participants within the Sichuan study, as well as more labelled events.}
        \label{fig:events_time}
    \end{subfigure}
    \hfill
    \begin{subfigure}[b]{0.48\textwidth}
        \centering
        \includegraphics[width=\textwidth]{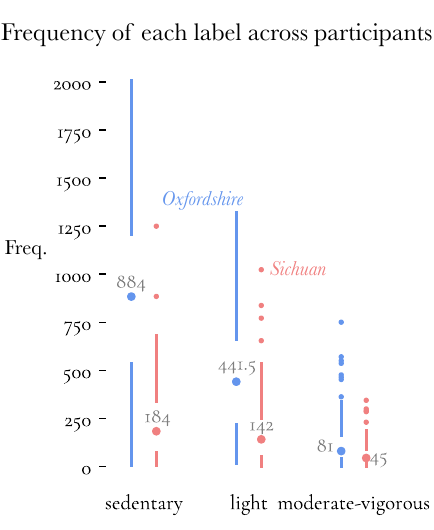}
        \caption{Quartile plot showing the imbalance in the prevalence of the activity intensity labels, and the relatively low number of instances of each label in the Sichuan study versus the Oxfordshire study.}
        \label{fig:label_imbalance}
    \end{subfigure}
    
    \vspace{0.1in}
    
    \begin{subfigure}[b]{\textwidth}
    \includegraphics[width=0.9\textwidth]{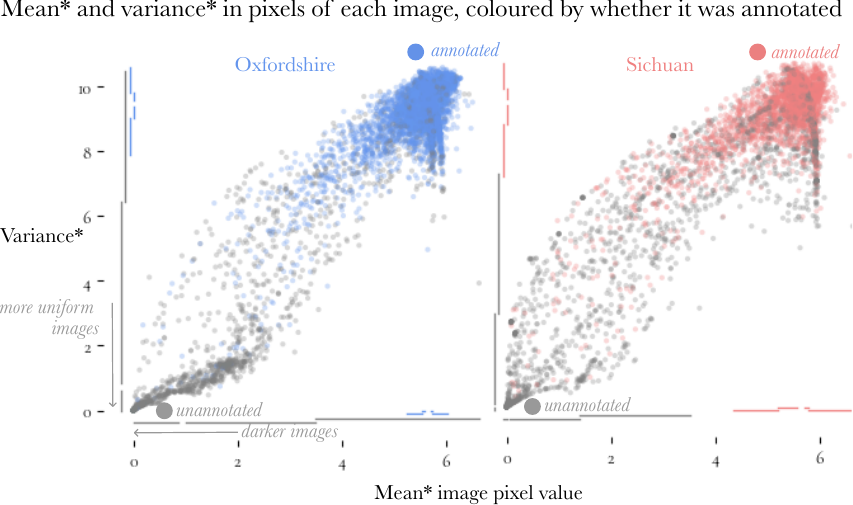}
    \caption{Scatter plot illustrating the relationship between unannotated images and images with low mean* and variance* pixel values. The mean* of the pixel values in each image was calculated as $\log(1+\sum_c \mu_c)$, where $\mu_c$ represents the mean of pixel values in an RGB image in channel $c$, and the variance* in the pixel values is an analogous transformation of the per channel variances. Intuitively, darker images will have lower mean pixel values, and images which are uniformly grey (or any other colour), will have no variance in their pixel values.}
    \label{fig:2dhist}
    \end{subfigure}

    \caption{Visualisation of the temporal sparsity of images, the label imbalance, and the large number of obscure images in the Oxfordshire and Sichuan validation study. The median participants day has 100 labelled events in the Oxfordshire study, versus 50 in the Sichuan study, with the much lower capture rate in this study potentially limiting the number of events that could be labelled. The majority of images were labelled as depicting sedentary activity.}
    \label{fig:data_supp_plot}
\end{figure}

Figure \ref{fig:events_time} shows the relationship between the median time between images and the number of labelled events per participant. The median time differences for the participants in the Oxfordshire study are clustered in 4 bands with the two most prominent clusters located around 20s, compared to the Sichuan study, whose participants are clustered in a band located at a median time of around 80s. There does not seem to be a strong relationship between these variables, since at fixed median time between images, we observe a large variation in the number of labelled events, though intuitively, at extremely low time intervals it is likely that many brief activities are missed, and it becomes impossible to accurately distinguish the timing of events. Figure \ref{fig:label_imbalance} shows quartile plots of the frequency of each label per participant. In addition to the class imbalance picked up in Table \ref{tab:demographics}, this shows the large range in the prevalence of the classes across participants. 

Finally, Figure \ref{fig:2dhist}, which is a scatter plot of images with the x-coordinate showing the mean pixel value of each image as a proxy for how dark it is, and the y-coordinate the variance in the pixel values as a proxy for how dynamic it is, illustrates the many obscure unlabelled images. From Table \ref{tab:demographics}, only 74\% of the images in the Oxfordshire study, and a much lower 34\% of the images in the Sichuan study were labelled with non-trivial labels. There were a few ways annotators expressed that they were unable to label images, including ``image dark/blurred/obscured'', ``camera taken off'', ``undefined''and ``unknown''. Table \ref{tab:uncodeable_comparison} shows the percentage of the images which could not be labelled for a particular reason. For completeness, which were simply not labelled. In the main text, we take labelled to mean an image has a non-trivial label. 
\begin{table}
\centering
\caption{Percentage of images labelled as uncodeable, unknown or undefined.}
\begin{tabular}{ r c c }
\hline
~ & Oxfordshire & Sichuan \\ \hline
\texttt{uncodeable;0002 image dark/blurred/obscured} & 16.40\%              & 56.98\%          \\
\texttt{uncodeable;0001 camera taken off}      & 1.68\%               & 6.91\%           \\ 
\texttt{undefined}                             & 0.17\%               & 0.04\%           \\ 
\texttt{<unknown>}                             & 0.01\%               & 0.00\%           \\ 
\hline
\end{tabular}
\label{tab:uncodeable_comparison}
\end{table}

\subsection{Implementation}
\label{app:implementation}

\begin{table}
    \centering
    \caption{Huggingface model IDs, number of parameters and size of each model.}
    \begin{tabular}{c c p{0.8in} }
    \hline
         Zero-shot models& Huggingface model ID   &No. parameters (millions)\\
         \hline
         CLIP & openai/clip-vit-large-patch14   &428\\
         BLIP2 & Salesforce/blip2-flan-t5-xl   &3 942\\
         LLaVA & llava-hf/llava-1.5-7b-hf   &7 063\\
         \hline
 Fine-tuned models&  &\\
         \hline
 ResNet-50 & IMAGENET1K\_V2  &25\\
 ViT (CLIP image encoder)& openai/clip-vit-large-patch14   &304\\
 \hline
    \end{tabular}

    \vspace{0.5em}
    \footnotesize
    \textit{Note:} For the ResNet, we used the torchvision ImageNet1K V2 checkpoint \citep{paszke2019pytorch}
    \label{tab:model_implement}.
\end{table}

Table \ref{tab:model_implement} gives the Hugging Face model IDs for the models used in this work, as well as the model sizes. Models weights were represented using 16-bit floating point precision (torch.float16), and were able to run on a single Tesla V100 with 32GiB of VRAM. Table \ref{tab:model_hparams} shows the hyperparameters tuned for each model. For the generative models, \verb|reword labels| controlled whether the text representations for sedentary behaviour, LIPA and MVPA were ``sedentary'', ``light'', ``MVPA'', or ````sedentary behavior'',  ``light physical activity'', and ``moderate-to-vigorous physical activity''. The set of prompts were too long to include in the table and are listed in the configuration files in the repository.
\begin{table}
    \centering
    \caption{Hyperparameters tuned for each model.}
    \begin{tabular}{c c}
        \hline 
        Hyperparameter & Values \\
        \hline
        \verb|mapping approach| & direct, via clean\\
        \verb|new tokens| & 5,10,20,40\\
        \verb|prompt| & ... \\
        \verb|reword labels| & true, false\\
        \hline
        \verb|batch size|  & 32, 64, 128, 256, 512\\
        \verb|finetune|& last layer, full model\\
        \verb|learning rate|  & $10^{-i}, i\sim U(1,5)$\\
        \verb|trivial augment|  & true, false\\
        \hline
         Zero-shot models&Hyperparameters tuned\\
         \hline
         CLIP & \verb|mapping approach|\\
         BLIP2 & \verb|mapping approach|, \verb|new tokens|, \verb|prompt|, \verb|reword labels|\\
         LLaVA & \verb|mapping approach|, \verb|new tokens|, \verb|prompt|, \verb|reword labels|\\
         \hline
 Fine-tuned models&\\
         \hline
 ResNet& \verb|finetune|, \verb|learning rate|, \verb|batch size|, \verb|trivial augment|\\
 ViT& \verb|finetune|, \verb|learning rate|, \verb|batch size|, \verb|trivial augment|\\
 ResNet-LSTM&\verb|learning rate|*\\
  \hline
    \end{tabular}

    \vspace{0.5em}
    \footnotesize
    \textit{Note:} We only tried three different learning rates for the ResNet-LSTM, $l \in \{10^{-3}, 10^{-4}, 10^{-5}\}$.
    \label{tab:model_hparams}
\end{table}

\subsection{Additional results}
\label{app:more_results}
\begin{figure}
    \centering
    \includegraphics[width=\linewidth]{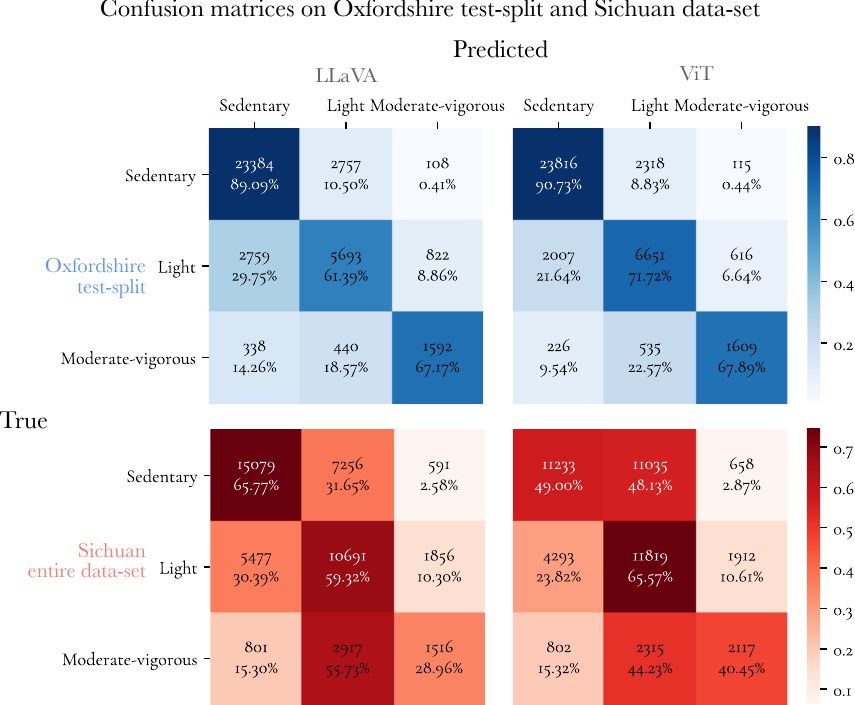}
    \caption{Confusion matrices showing the disagreements between the human and model predictions, for LLaVA and ViT, particularly on the Sichuan study. The percentages (and colours) are normalised based on the total number of ``true'' instances of each label.}
    \label{fig:conf_mat}
\end{figure}

Figure \ref{fig:conf_mat} shows confusion matrices for the best checkpoint for LLaVA and ViT. These confusion matrices ignore variation in performance at the participant level, though facilitate comparisons to work by \cite{keadle2024using}. We include the converted confusion matrix from this work in Table \ref{tab:keadle_conf_mat}.

\begin{table}[ht]
    \centering
    \caption{Confusion matrix from Table 3 of \cite{keadle2024using}, showing the performance of XGBoost \citep{chen2016xgboost} based on features from AlphaPose \citep{fang2022alphapose} with the rows and columns related to moderate and vigorous physical activity combined.}
    \begin{tabular}{c|c c c}
    \hline
         True / Predicted & Sedentary & Light & MVPA \\
         \hline
         Sedentary & 13 259 & 4 915 & 345\\
         LIPA & 197 & 939 & 129 \\ 
         MVPA & 1255 & 2427 & 6594 \\ 
         \hline
    \end{tabular}
    \label{tab:keadle_conf_mat}
\end{table}

\begin{table}
    \centering
    \caption{Examples of the raw captions produced by different prompts and models, and the labels they were mapped to. Some of these captions were first mapped via one of the clean labels associated with each coarse label.}
    \begin{tabular}{p{1.8in}p{1.1in}p{0.8in}cl}
    \hline
          Caption $\rightarrow$& Mapped via $\rightarrow$&Mapped to& Sim. &\\
         \hline
            a woman sitting in a chair and talking to a woman&  sitting meeting or talking with others&  sedentary behaviour& 0.47 &\ding{51} \\
 a fence and a yard& mowing lawn& MVPA& 0.41 &? \\
           a woman playing with a frisbee&  bowling&  LIPA & 0.32 &\ding{55} \\
         \hline
    \end{tabular}
    \label{tab:qualitative_ex}
\end{table}

Sometimes, the captions produced by the generative models were mapped to labels which did not mean the same thing as the produced caption. In Table \ref{tab:qualitative_ex}, we give examples of the produced captions, the label they were ultimately mapped to, possibly via an intermediate clean label, as well as the similarity score from the sentence embedding model.

\subsection{N=1 human performance from single images}
\label{app:human_performance}
To estimate human performance for labelling activity intensity from single images, one of the authors (Abram Schönfeldt) manually labelled $\geq$ 500 images from participants in the test splits from the Oxfordshire (25 participants) and Sichuan (13 participants) validation studies. The images were sampled uniformly at random and presented without temporal context, which is not how these data-sets were originally labelled, though reflects the information seen by the models. The median $\kappa$ (1st, 3rd quartile) on the Oxfordshire test-split was 0.636 (0.457, 0.722), and 0.572 (0.464, 0.610) on the Sichuan study. Though limited by small amount of labelled data, and single annotator, these results suggest that the current model performance might actually be similar to human performance.

\end{document}